\documentclass[sigconf]{acmart}
\usepackage{subcaption}
\usepackage[british]{babel}
\usepackage{makecell}
\usepackage{multirow}

\AtBeginDocument{%
  \providecommand\BibTeX{{%
    \normalfont B\kern-0.5em{\scshape i\kern-0.25em b}\kern-0.8em\TeX}}}

\copyrightyear{2023}
\acmYear{2023}
\setcopyright{rightsretained}
\acmConference[CIKM '23]{Proceedings of the 32nd ACM International
Conference on Information and Knowledge Management}{October 21--25,
2023}{Birmingham, United Kingdom}
\acmBooktitle{Proceedings of the 32nd ACM International Conference on
Information and Knowledge Management (CIKM '23), October 21--25, 2023,
Birmingham, United Kingdom}
\acmDOI{10.1145/3583780.3615241}
\acmISBN{979-8-4007-0124-5/23/10}

\begin{document}

\title{KGrEaT: A Framework to Evaluate\\Knowledge Graphs via Downstream Tasks}
\renewcommand{\shorttitle}{KGrEaT: A Framework to Evaluate Knowledge Graphs via Downstream Tasks}

\author{Nicolas Heist}
\authornote{Both authors contributed equally to this research.}
\email{nico@informatik.uni-mannheim.de}
\orcid{0000-0002-4354-9138}
\author{Sven Hertling}
\authornotemark[1]
\email{sven@informatik.uni-mannheim.de}
\orcid{0000-0003-0333-5888}
\author{Heiko Paulheim}
\email{heiko@informatik.uni-mannheim.de}
\orcid{0000-0003-4386-8195}
\affiliation{%
  \institution{Data and Web Science Group, University of Mannheim}
  \city{Mannheim}
  \country{Germany}
}

\begin{abstract}
In recent years, countless research papers have addressed the topics of knowledge graph creation, extension, or completion in order to create knowledge graphs that are larger, more correct, or more diverse. This research is typically motivated by the argumentation that using such enhanced knowledge graphs to solve downstream tasks will improve performance. Nonetheless, this is hardly ever evaluated. Instead, the predominant evaluation metrics - aiming at correctness and completeness - are undoubtedly valuable but fail to capture the complete picture, i.e., how useful the created or enhanced knowledge graph actually is. Further, the accessibility of such a knowledge graph is rarely considered (e.g., whether it contains expressive labels, descriptions, and sufficient context information to link textual mentions to the entities of the knowledge graph). To better judge how well knowledge graphs perform on actual tasks, we present KGrEaT -- a framework to estimate the quality of knowledge graphs via actual downstream tasks like classification, clustering, or recommendation. Instead of comparing different methods of processing knowledge graphs with respect to a single task, the purpose of KGrEaT is to compare various knowledge graphs as such by evaluating them on a fixed task setup. The framework takes a knowledge graph as input, automatically maps it to the datasets to be evaluated on, and computes performance metrics for the defined tasks. It is built in a modular way to be easily extendable with additional tasks and datasets.
\end{abstract}




\maketitle

\section{Introduction}
\subsection{Motivation}
Knowledge graphs (KGs) have emerged as a powerful tool for organizing and representing structured knowledge in a machine-readable format. Starting with Google’s announcement of the Google Knowledge Graph in 2012\footnote{\url{https://blog.google/products/search/introducing-knowledge-graph-things-not/}}, research articles have extensively explored the creation \cite{dbpedia,wikidata,yago3}, extension \cite{heist2018language,heist2023nastylinker}, refinement \cite{paulheim2017knowledge}, and completion \cite{akrami2020realistic} of KGs, with the aim of producing larger, more accurate, and more diverse graphs. These efforts are driven by the belief that leveraging enhanced KGs can lead to improved performance in downstream tasks. However, comparative evaluations of different KGs w.r.t. their utility for such tasks are rarely conducted.

In the literature, the vast majority of studies concerned with the evaluation of KGs have focused on intrinsic metrics that are working exclusively with the triples of a graph. Several works introduce quality metrics like accuracy, consistency, or trustworthiness and propose ways to determine them quantitatively \cite{kgQuality,ban2022quality,kgaccurarcy,huaman2021towards,zaveri2012quality}. Färber et al. \cite{farber2018knowledge} and Heist et al. \cite{KGoverview} compare KGs with respect to size, complexity, coverage, and overlap. Additionally, they provide guidelines on which KG to select for a given problem.

Another line of work computes extrinsic task-based metrics to evaluate KG embedding approaches. They use a fixed input KG with a fixed evaluation setup while varying only the embedding approach. Frameworks like GEval~\cite{pellegrino2020geval} or kgbench~\cite{bloem2021kgbench} use data mining tasks like classification or regression for the evaluation, others, like Ali et al. \cite{bringinglight} evaluate primarily on link prediction tasks.

\subsection{Contributions}
To address the evaluation gap of extrinsic metrics for KGs, we propose a framework called KGrEaT (\textbf{K}nowledge \textbf{Gr}aph \textbf{E}v\textbf{a}luation via Downstream \textbf{T}asks).\footnote{\url{https://github.com/dwslab/kgreat}} KGrEaT aims to provide a comprehensive assessment of KGs by evaluating them on multiple kinds of tasks like classification, regression, or recommendation. The evaluation results (e.g., the accuracy of a classification model trained with the KG as background knowledge) serve as extrinsic task-based quality metrics for the KG. By defining a fixed evaluation setup in the framework and applying it to multiple KGs, it is possible to isolate the effect of every single KG and compare how useful they are for solving different kinds of tasks. KGrEaT is built in a modular way to be open for extensions from the community like additional tasks or datasets.

Overall, the contributions of this paper are as follows:
\begin{itemize}
\item With KGrEaT, we present a framework to judge the utility of KGs using extrinsic task-based metrics (Section~\ref{sec:framework}).
\item In our experiments, we demonstrate the capabilities of the framework in an evaluation and comparison of several well-known cross-domain KGs (Section~\ref{sec:experiments}).
\end{itemize}

\begin{figure*}[ht]
  \centering
  \includegraphics[width=.9\textwidth]{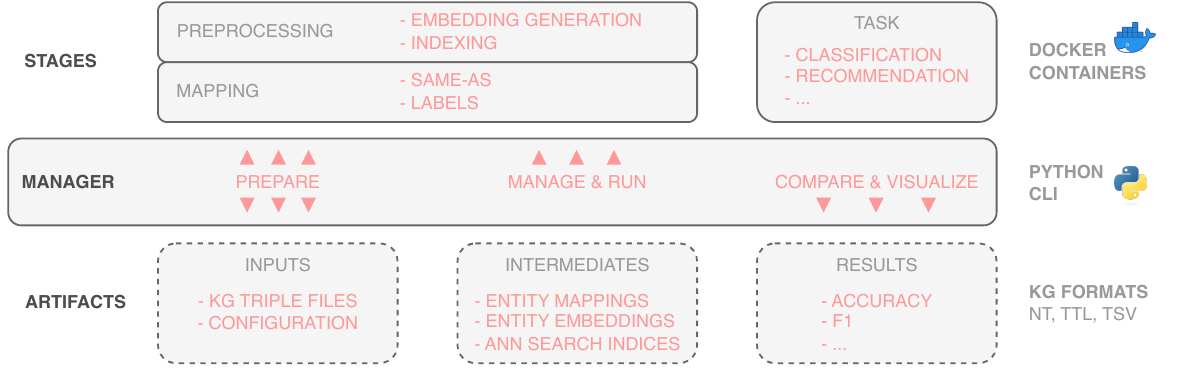}
  \caption{An overview of the KGrEaT framework.}
  \label{fig:approach-overview}
\end{figure*}

\section{Framework}
\label{sec:framework}
\subsection{Purpose and Limitations}
KGrEaT is a framework built to evaluate the performance impact of KGs on multiple downstream tasks. To that end, the framework implements various algorithms to solve tasks like classification, regression, or recommendation of entities. The impact of a given KG is measured by using its information as background knowledge for solving the tasks. To compare the performance of different KGs on downstream tasks, we use a fixed experimental setup with the KG as the only variable. The performance indicators may be used to make an informed decision when picking a KG for a given task. Further, they can be used to compare the performance of different versions of a single KG (e.g., during construction or during its life cycle).

The implemented algorithms are not necessarily state-of-the-art because the primary objective is not to measure how well a task can be solved with a given KG in absolute numbers, but rather in comparison to other KGs or different versions of the same KG. Hence, the absolute numbers of the results only have limited expressive power. However, the framework tries to reduce the bias in the results by averaging over multiple preprocessing methods, datasets, and algorithms.

KGrEaT maps the entities of the KG automatically to the entities of the dataset using a set of configurable mappers. Undoubtedly, the quality of this mapping influences the results generated by the framework. But as the mapping procedure is applied similarly to all evaluated KGs, the mapping quality is mainly influenced by the accessibility of the graph (i.e., whether it provides sufficient context information like labels or descriptions for its entities).
To reduce the influence of the mapping strategy on the overall results, the framework provides a way to run experiments with multiple mapping approaches (and possibly average over them).

\begin{table*}[ht]
\caption{Implemented tasks together with their algorithms, datasets, and evaluation metrics. }
\label{tab:tasks}
\resizebox{\textwidth}{!}{%
\begin{tabular}{llll}
\toprule
\textbf{Task Type} & \textbf{Datasets} & \textbf{Algorithms} & \textbf{Evaluation Metrics} \\
\midrule
Classification & \makecell[l]{AAUP, Cities, Forbes, MillionSongDataset\\ MetacriticAlbums, MetacriticMovies, ComicCharacters} & Naive Bayes, KNN, SVM & Accuracy $\uparrow$ \\ \midrule
Regression & \makecell[l]{AAUP, Cities, Forbes,\\ MetacriticAlbums, MetacriticMovies} & \makecell[l]{Linear Regression, KNN,\\ Decision Tree} & RMSE $\downarrow$ \\ \midrule
Clustering & \makecell[l]{Cities2000AndCountries, CitiesAndCountries, Teams,\\ CitiesMoviesAlbumsCompaniesUni, ComicCharacters} & \makecell[l]{DBSCAN, KMeans,\\ Agglomerative Clustering} & ARI $\uparrow$, NMI $\uparrow$, Accuracy $\uparrow$ \\ \midrule
Document Similarity & LP50 & Cosine Similarity & \makecell[l]{Spearman $\uparrow$, Pearson $\uparrow$} \\ \midrule
Entity Relatedness & KORE & Cosine Similarity & Kendall's Tau $\uparrow$ \\ \midrule
Semantic Analogies & \makecell[l]{AllCapitalCountryEntities, CapitalCountryEntities,\\ CityStateEntities, CurrencyEntities} & Cosine Similarity & Accuracy $\uparrow$ \\ \midrule
Recommendation & MovieLens, LastFm, LibraryThing & Item-Similarity recommender & F1 $\uparrow$ \\
\bottomrule
\end{tabular}
}
\end{table*}

\subsection{Design}
The framework is designed in a modular way (c.f. Figure~\ref{fig:approach-overview}), making it easy to add additional preprocessing steps, mappers, or tasks. Every step of a stage is implemented as an isolated docker container\footnote{\url{https://www.docker.com}} with its own environment so that additions can be made without any constraints on the programming language. Another advantage of the container-based architecture is the easy distribution of containers via a container hub, eliminating the need for users to build the framework on their own machines.

The manager is responsible for making necessary preparations (e.g., downloading the input data or gathering entities to be mapped), executing the stages (fetching and running containers of the steps), and visualizing the results (e.g., comparing KG performance on various aggregation levels). The \texttt{Preprocessing} and \texttt{Mapping} stages can be executed in parallel, and the results are then used to execute the \texttt{Task} stage. The whole process can be steered via a command line interface (CLI).

The only input to the evaluation process is the KG in the form of RDF files as well as a configuration. The latter defines how the stages should be run (i.e., which steps to execute in which order). Further, every step can be configured in depth to supply relevant hyper-parameters. For example, one can configure how the KG should be mapped to the datasets (e.g., via matching labels) and define an acceptable similarity value for a match.

In the following, we provide details of the three main stages that are executed when running an evaluation of a KG.

\subsection{Preprocessing Stage}
The \texttt{Preprocessing} stage creates all pre-computable artifacts that are needed in the subsequent \texttt{Task} stage (e.g., intermediate representations or statistics of the KG). So far, this stage comprises the computation of embeddings ($TransE$~\cite{transe}, $TransR$~\cite{transr}, $DistMult$~\cite{distmult}, $RESCAL$~\cite{rescal}, $ComplEx$~\cite{complex} via the DGL-KE framework \cite{zheng2020dgl}, and RDF2vec~\cite{ristoski2016rdf2vec} via the jRDF2vec framework~\cite{portisch2020rdf2vec}). Further, it supports the generation of indices for Approximate Nearest Neighbor (ANN) search (via the hnswlib library \cite{malkov2018efficient}).

\subsection{Mapping Stage}
In the \texttt{Mapping} stage, the entities of the KG are automatically mapped to the entities in the datasets. So far, a \texttt{Same-As} mapper and a \texttt{Label} mapper are implemented. The former uses the same-as links of a KG to map its entities to those of the datasets. A dataset may provide URIs for an entity (e.g., from well-known KGs like DBpedia or Wikidata), but it has to provide at least one label. This label is used by the \texttt{Label} mapper to find a corresponding entity in the KG. It uses the RapidFuzz library\footnote{\url{https://github.com/maxbachmann/RapidFuzz}} to estimate the similarity of labels via token-based edit distance. Mappers are composable, i.e., they can be executed in sequence. For example, entities are first mapped via same-as links where available, and the remaining entities are mapped via label similarity.

\subsection{Task Stage}
In the \texttt{Task} stage, the task types are executed for all combinations of datasets and algorithms. Table~\ref{tab:tasks} gives an overview of all possible constellations. In total, KGrEaT contains 26 tasks (i.e., combinations of task types and datasets) that are run with one or more algorithms. Additionally, the algorithms are executed with multiple hyperparameter settings. How the individual tasks use the KG information is dependent on the task and the implemented algorithm. Generally, the tasks \texttt{Classification}, \texttt{Regression}, and \texttt{Clustering} use embeddings of the KG's entities as features of the models, and the remaining tasks use the distance between the entity embeddings to find related entities.

Several datasets are taken from Ristoski et al. \cite{ristoski2016collection} and from the GEval framework \cite{pellegrino2020geval}. The \texttt{Recommendation} datasets MovieLens \cite{harper2015movielens}, LastFm\footnote{\url{http://www.lastfm.com}}, and Library\-Thing \cite{zhao2015improving} are preprocessed as recommended by Di Noia et al. \cite{noia2016sprank} with the exception of using all entities instead of only those for which a mapping to DBpedia exists. For detailed statistics of all datasets, please refer to the respective publications and the information in the framework.

Every task type comes with suitable evaluation metrics that are computed for every constellation. As some KGs might not contain matches for all entities in the dataset and it would not be fair to compute metrics only over known entities (and discard unknown entities) or only over all entities, the framework reports metrics for both scenarios. Finally, the results can be aggregated over various levels (e.g., over embeddings, algorithms, and datasets) to produce metrics with a reduced bias.

\begin{table*}
	\caption{Evaluation results of the KGs aggregated by task type and metric. The results of the KGs are given for the dimensions PK (precision-oriented, known entities), PA (precision-oriented, all entities), and RA (recall-oriented, all entities).}
	\label{tab:results}
	\resizebox{\textwidth}{!}{%
		\begin{tabular}{ll|ccc|ccc|ccc|cc|ccc|ccc}
			\toprule
			Task Type      & Metric        &         \multicolumn{3}{c}{DBpedia2016}          &         \multicolumn{3}{c}{DBpedia2022}          &             \multicolumn{3}{c}{YAGO}             &  \multicolumn{2}{c}{Wikidata}   & \multicolumn{3}{c}{CaLiGraph} & \multicolumn{3}{c}{DbkWik} \\
			               &               &       PK       &       PA       &       RA        &       PK       &       PA       &       RA        &       PK       &       PA       &       RA        &       PK       &       PA       &  PK   &   PA   &      RA       &  PK   &  PA   &     RA      \\ \midrule
			Classification & Accuracy      &         \textbf{0.576} &  \textbf{0.476} &  0.434 &         0.561 &  0.433 &  0.394 &   0.559 &  0.464 &  0.527 &      0.559 &  0.366 &       0.565 &  0.467 &  \textbf{0.529} &    0.539 &  0.467 &  0.501 \\
			Regression     & RMSE          &        0.693 & \textbf{1.271} & 1.320 &        0.688 & 1.287 & 1.284 &  0.706 & 1.331 & \textbf{0.720} &     \textbf{0.684} & 1.300 &      0.734 & 1.321 & 0.855 &   0.718 & 1.287 &    1.350 \\
			Clustering     & ARI           &         0.149 &  \textbf{0.218} &  \textbf{0.217} &         0.115 &  0.176 &  0.187 &   \textbf{0.240} &  0.216 &  0.213 &      0.055 &  0.076 &       0.138 &  0.139 &  0.125 &    0.192 &  0.193 &    0.199 \\
			               & NMI           &         \textbf{0.303} &  \textbf{0.248} &  \textbf{0.243} &         0.272 &  0.213 &  0.216 &   0.278 &  0.245 &  0.213 &      0.178 &  0.164 &       0.169 &  0.190 &  0.132 &    0.187 &  0.200 &    0.191 \\
			               & Accuracy      &         \textbf{0.762} &  0.614 &  0.633 &         0.740 &  0.556 &  0.577 &   0.754 &  0.560 &  \textbf{0.699} &      0.678 &  0.410 &       0.708 &  0.547 &  0.660 &    0.691 &  \textbf{0.681} &    0.689 \\
			Doc. Sim.      & Spearman      &         0.207 &  0.207 &  0.207 &         \textbf{0.226} &  \textbf{0.226} &  \textbf{0.226} &   0.165 &  0.165 &  0.160 &      0.131 &  0.165 &       0.203 &  0.203 &  0.153 &    0.200 &  0.200 &    0.214 \\
			               & Pearson       &         0.294 &  0.294 &  0.294 &         \textbf{0.306} &  \textbf{0.306} &  \textbf{0.306} &   0.235 &  0.235 &  0.233 &      0.241 &  0.069 &       0.274 &  0.274 &  0.226 &    0.274 &  0.274 &    0.283 \\
			               & Harm. Mean    &         0.241 &  0.241 &  0.241 &         \textbf{0.257} &  \textbf{0.257} &  \textbf{0.257} &   0.184 &  0.184 &  0.191 &      0.172 &  0.103 &       0.230 &  0.230 &  0.180 &    0.164 &  0.164 &    0.241 \\
			Ent. Rel.      & Kendall's Tau &         0.135 &  0.104 &  0.109 &         0.179 &  0.108 &  \textbf{0.119} &   0.012 &  0.015 &  0.008 &      \textbf{0.203} &  0.071 &       0.086 &  0.056 &  0.078 &    0.134 &  \textbf{0.119} &    0.117 \\
                Sem. Analogies      & Accuracy   &         0.253 &  0.246 &  \textbf{0.261} &         \textbf{0.265} &  \textbf{0.249} &  0.247 &   0.221 &  0.219 &  0.214 &      0.001 &  0.000 &       0.219 &  0.187 &  0.198 &    0.215 &  0.212 &    0.206 \\
			Recommend.     & F1            &         0.015 &  \textbf{0.011} &  \textbf{0.011} &         0.014 &  \textbf{0.011} &  0.010 &   0.008 &  0.006 &    0.006	 &      \textbf{0.021} &  0.006 &       0.013 &  0.009 &  0.009 &    0.011 &  0.010 &    0.009 \\
            \bottomrule
		\end{tabular}
	}
\end{table*}

\section{Experiments}
\label{sec:experiments}
To show the capabilities of KGrEaT, we conduct experiments over multiple large cross-domain KGs and analyze how well they perform on the implemented downstream tasks. We first give an overview of the evaluated KGs, then define the experimental setup, and finally discuss the results.

\subsection{Experimental Setup}
\subsubsection{Knowledge Graphs}
We use the following KGs in our experiments:
\begin{enumerate}
    \item \textit{DBpedia} \cite{dbpedia}: Dumps from 2016-10 and 2022-09\footnote{Using multiple versions allows us to compare not only between different KGs but also between different versions (here: with respect to time) of the same KG.}
    \item \textit{YAGO} \cite{yago3}: Version 3
    \item \textit{Wikidata} \cite{wikidata}: Dump from 2023-06-07
    \item \textit{CaLiGraph} \cite{heist2019uncovering, heist2021caligraph}: Version 3.1.1
    \item \textit{DBkWik} \cite{dbkwik}: Version DBkWik++\footnote{Combined with DBpedia to also include the well-known entities of Wikipedia}
\end{enumerate}

\subsubsection{Mapping}
We first map the KGs with the \texttt{Same-As} mapper where applicable. Then we apply two variants of the \texttt{Label} mapper: One with a similarity threshold of 1.0 for high-precision matches and one with a threshold of 0.7 for high recall. For the former, we compute metrics for known entities (\textbf{P}recision \textbf{K}nown - PK) and for all entities (\textbf{P}recision \textbf{A}ll - PA); for the latter, being recall-oriented, we report the metrics only for all entities (\textbf{R}ecall \textbf{A}ll - RA).

\subsubsection{Embeddings}
To reduce the influence of the different embedding approaches on the overall results, all experiments are executed with four embedding types ($TransE$, $DistMult$, $ComplEx$, and $RDF2vec$). For Wikidata, we could not compute all these embeddings due to the amount of computational resources necessary. Instead, we use pre-computed $TransE$ embeddings\footnote{\url{https://torchbiggraph.readthedocs.io/en/latest/pretrained_embeddings.html}} with a comparable training configuration.

\subsubsection{Hardware}
All experiments are executed on NVIDIA RTX 2080 Ti graphic cards and Intel Xeon E5 processors (2.6GHz). On average, a full evaluation of a single KG takes roughly 30 hours with 20 hours of embedding computation, 4 hours of mapping, and 6 hours of task execution.

\subsection{Results and Discussion}
Table~\ref{tab:results} shows the final results of our evaluation for the three scenarios $PK$, $PA$, and $RA$. The results are averaged after aggregating over all embeddings, datasets, and algorithms. The complete results of the experiments are publicly available.\footnote{\url{https://doi.org/10.5281/zenodo.8050446}}

For \texttt{Classification}, DBpedia2016 shows the best results in the precision setting, while CaLiGraph and YAGO achieve the best results in the recall setting. For \texttt{Regression}, both DBpedia versions and Wikidata perform well in the precision setup, while again YAGO and CaLiGraph achieve the best results in the recall setting. The \texttt{Clustering} task is solved best by DBpedia2016, YAGO, and DbkWik. For \texttt{Document Similarity}, version 2022 of DBpedia is the clear winner. For the \texttt{Entity Relatedness} task, using DBpedia2022, Wikidata, or DbkWik as background knowledge produces the best results. \texttt{Recommendation} is solved best using DBpedia or Wikidata, \texttt{Semantic Analogies} is also solved best by DBpedia.

In general, DBpedia dominates the results to a large extent which may be explained by the fact that some of the datasets used in the framework have been derived from the 2015 version of DBpedia. This might also explain that there is no clear advantage of the 2022 version of DBpedia over the older 2016 version. However, both versions of DBpedia perform strongly on the \texttt{Recommendation} task which has no direct relation to DBpedia or even Wikipedia.

Our assumption that the KGs with more entities (YAGO, Wikidata, CaLiGraph, and DBkWik) will have an advantage, especially in the \texttt{Recommendation} tasks, did only partially prove to be true. However, they have shown strong performances, especially in recall-oriented settings. A reason for this unsteady performance may lie in the increased complexity of training expressive embeddings for large KGs. In the future, we want to explore this further by running evaluations not only with multiple types of embeddings but also with multiple embedding configurations (e.g., number of trained epochs). Another interesting direction to explore is whether combining two KGs (e.g., by concatenating their entity vectors) yields improved results \cite{thoma2017towards}.

\section{Conclusion and Outlook}
We presented KGrEaT, a framework for evaluating the performance of KGs on multiple downstream tasks. In our experiments, we found that, depending on the task, the performance of the KGs varies enormously. To judge the quality of a KG in its completeness, extrinsic evaluation metrics provided by KGrEaT can serve as a valuable addition to the established intrinsic evaluation criteria.

In the future, we want to improve the framework in various ways, e.g., by providing more embedding methods such as RDF2Vec \cite{rdf2vec} as well as more tasks like KG Question Answering \cite{kgqadataset}.

Further, we plan to include a more comprehensive mapper that uses all information of an entity (such as comments and relations to other entities). To that end, we transform the entities of the datasets into a small KG which is then mapped to the entities of the KG under evaluation. In such a case, systems participating in the Ontology Alignment Evaluation Initiative (OAEI) \cite{oaei2022} may prove useful.

To open the framework for users unfamiliar with programming and docker, we will introduce a graphical user interface, allowing them to analyze KGs in a faster and more intuitive way.

\bibliographystyle{ACM-Reference-Format}
\balance
\bibliography{sample-base}

\end{document}